\documentclass[sigconf]{acmart}
\usepackage{multirow}
\usepackage{booktabs}

\usepackage{colortbl}
\usepackage{xcolor}
\usepackage{graphicx}
\usepackage[table]{xcolor}

\renewcommand\footnotetextcopyrightpermission[1]{} 

\usepackage{fancyhdr}
\usepackage{pifont}
\usepackage{stfloats} 

\usepackage{graphicx}
\usepackage{textcomp}
\usepackage{xcolor}
\usepackage{algorithm}
\usepackage[noend]{algpseudocode}
\algrenewcommand\algorithmicrequire{\textbf{Input:}}
\algrenewcommand\algorithmicensure{\textbf{Output:}}
\algrenewcommand\algorithmiccomment[1]{\hfill$\triangleright$~#1}
\algrenewcommand\alglinenumber[1]{\footnotesize #1:}

\newcommand{\cmt}[1]{\textcolor{gray!55!black}{\Statex\textbf{\texttt{/*}}\textbf{\texttt{  #1}}\hfill\textbf{\texttt{*/}}}}

\usepackage{booktabs}

\setcitestyle{numbers,sort&compress}

\AtBeginDocument{%
  }

\begin{document}

\title{Crystal-KV: Efficient KV Cache Management for Chain-of-Thought LLMs via Answer-First Principle}


\author{Zihan Wang}
\affiliation{%
  \institution{School of Computer Science and Technology, University of Science and Technology of China}
  \city{Hefei}
  \country{China}
  }
\email{wangzh196@mail.ustc.edu.cn}

\author{Cheng Tang}
\affiliation{%
  \institution{School of Computer Science and Technology, University of Science and Technology of China}
  \city{Hefei}
  \country{China}
  }
\email{sisyphustc@mail.ustc.edu.cn}

\author{Lei Gong}
\affiliation{%
  \institution{School of Computer Science and Technology, University of Science and Technology of China}
  \city{Hefei}
  \country{China}
  }
\email{leigong0203@ustc.edu.cn}

\author{Cheng Li}
\affiliation{%
  \institution{University of Science and Technology of China; Institute of Artificial Intelligence, Hefei Comprehensive National Science Center}
  \city{Hefei}
  \country{China}
  }
\email{chengli7@ustc.edu.cn}

\author{Chao Wang}
\affiliation{%
  \institution{School of Computer Science and Technology, University of Science and Technology of China}
  \city{Hefei}
  \country{China}
  }
\email{cswang@ustc.edu.cn}
  
\author{Teng Wang}
\affiliation{%
  \institution{Suzhou Institute for Advanced Research, University of Science and Technology of China}
  \city{Suzhou}
  \country{China}
  }
\email{wangt635@ustc.edu.cn}

\author{Wenqi Lou}
\affiliation{%
  \institution{Suzhou Institute for Advanced Research, University of Science and Technology of China}
  \city{Suzhou}
  \country{China}
  }
\email{louwenqi@ustc.edu.cn}

\author{Xuehai Zhou}
\affiliation{%
  \institution{School of Computer Science and Technology, University of Science and Technology of China}
  \city{Hefei}
  \country{China}
  }
\email{xhzhou@ustc.edu.cn }







\begin{abstract}
Chain-of-Thought (CoT) reasoning in large language models (LLMs) significantly improves accuracy on complex tasks, yet incurs excessive memory overhead due to the long think-stage sequences stored in the Key-Value (KV) cache. Unlike traditional generation tasks where all tokens are uniformly important, CoT emphasizes the final answer, rendering conventional KV compression strategies ineffective. In this paper, we present Crystal-KV, an efficient KV cache management framework tailored for CoT reasoning. Our key insight is the answer-first principle. By mapping answer preferences into think-stage attention map, we distinguish between SlipKV, which mainly maintains the reasoning flow but may occasionally introduce misleading context, and CrystalKV, which truly contributes to the correctness of the final answer. Next, we propose an attention-based Least Recently Frequently Used algorithm. It precisely identifies when a SlipKV entry's utility expires and evicts it, retaining CrystalKV without disrupting reasoning flow. Finally, we introduce an adaptive cache budget allocation algorithm. Based on the dynamic proportion of CrystalKV, it estimates the importance of each layer/head and adjusts the KV cache budget during inference, amplifying critical components to improve budget utilization. Results show that Crystal-KV achieves state-of-the-art KV cache compression, significantly improves throughput, and enables faster response time, while maintaining, or even improving, answer accuracy for CoT reasoning. \textbf{\texttt{The Code will be open-sourced at github.com/xxx}}.
\end{abstract}

\keywords{KV Cache Compression, Efficient CoT Reasoning, LLM Inference}

\maketitle

\thispagestyle{empty}  
\pagestyle{empty}      

\section{Introduction}
\label{intro}
\begin{figure}[t]
  \centering
  \includegraphics[width=\linewidth]{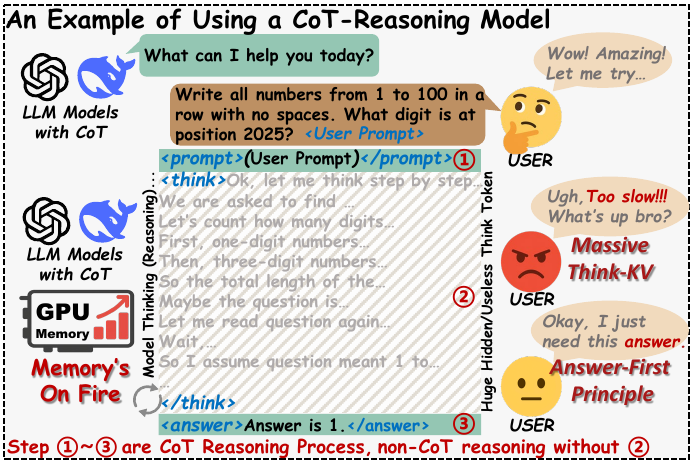}
  \caption{The Workflow of Chain-of-Thought Reasoning}
  \Description{High-level architecture of Crystal-KV showing KV cache partitioning and dynamic reallocation across layers and heads.}
  \label{fig:intro_fig}
\end{figure}

Chain-of-Thought (CoT) reasoning has been widely adopted in large language models (LLMs) to solve complex tasks such as mathematics and programming, with notable success in models like ChatGPT 5 (Thinking)~\cite{openai2025gpt5}, 
DeepSeek R1~\cite{guo2025deepseek}, 
Qwen3~\cite{yang2025qwen3}, 
and Gemini 2.5~\cite{comanici2025gemini}. The key of CoT is to insert a think stage before generating the final answer~\cite{wei2022chain}. As shown in Fig.~\ref{fig:intro_fig}, the user first submits a prompt. The LLM then enters a think stage (i.e., the CoT stage), where it generates a massive number of tokens for knowledge expansion and logical deduction. Finally, the LLM outputs the final answer. \textit{Importantly, the intermediate think stage is often hidden or useless to users, while what users care about is the final answer}. Despite CoT's potential for accuracy, it incurs substantial memory overhead, as massive think-stage tokens correspond to a large Key-Value (KV) cache~\cite{gao2024cost}. For example, DeepSeek-R1-Distill-Qwen-14B may consume 8K tokens to solve a complex coding problem, requiring 10 GB of memory for KV storage, hundreds of times larger than the prompt and answer stage KV. This dramatically increases memory usage and computational cost, while also slowing down response time for users. Therefore, compressing the think-stage KV cache is essential to enable efficient deployment of CoT reasoning.

The key to KV compression is dynamically evicting redundant KV cache entries during inference while preserving accuracy~\cite{cai2025r,xiao2024efficient}. Although many compression methods exist, most are limited to normal long-content generation (LCG) tasks, such as dialogue systems, and fail to generalize to reasoning tasks with CoT. The fundamental reason lies in the generation objective shift: from producing \textit{\textbf{ALL}} output tokens to focusing \textit{\textbf{ONLY}} on the final answer. In LCG tasks, since every token is relevant to the user, these compression methods fairly maintain the quality of each token generation. As a result, by approximating attention scores in the near future, they greedily evict KV entries that are lowly attended by the next few tokens. In contrast, in reasoning tasks with CoT, users only care about the final answer. Therefore, \textbf{uniform token treatment} and \textbf{answer-first prioritization} are inherently conflicting optimization goals. Even worse, the shortsighted greed induced by the uniform treatment significantly harms the quality of answer-stage tokens, which are generated at the end of the sequence. While some work explores KV compression for CoT~\cite{hu-etal-2025-raas,cai2025r}, they overlook the decisive role of the final answer and treat the think stage as a special case of LCG.

Our key insight is the answer-first principle: retain only those KV entries that contribute to the final correct answer during the think stage, without disrupting the reasoning flow. However, three challenges arise: (i) Temporal lag: Since the answer stage occurs entirely after the think stage, it is infeasible to reliably assess a think-stage KV entry’s importance to the final answer based on attention score approximation. This raises a fundamental question: \textit{Do the think-stage KV entries that contribute to the correct answer exhibit any unified attention pattern during the think stage?} If so, it could serve as the basis for early identification of critical KV entries before the answer stage begins. (ii) Dynamism: CoT reasoning is a dynamic logical deduction, where important think-stage KV entries may emerge at any time. This requires not only maintaining the evolving reasoning flow, but also accurately retaining critical KV entries in real time. (iii) The combined effects of temporal lag and dynamism push KV-cache budget allocation into a dilemma: either wasting cache space or missing critical KV entries.

To address these challenges, we propose Crystal-KV, an efficient KV cache management framework for CoT reasoning in LLM inference. (i) By mapping answer preferences into the think-stage attention map, we reveal two unified attention patterns. The KV entries truly contributing to the correct answer are intermittently, rather than continuously, attended until the end of reasoning. We refer to these as CrystalKV. Meanwhile, others exhibit a streaming attention pattern. Though helpful in maintaining reasoning flow, these tokens may occasionally introduce misleading information for the final answer, and are thus referred to as SlipKV. Excitingly, our experiments show that focusing more attention on CrystalKV during answer stage can even turn previously wrong answers into correct ones. (ii) Based on unified attention patterns, we propose an attention-based Least Recently Frequently Used (LRFU) algorithm. It precisely identifies when a SlipKV entry's utility expires and performs eviction, ensuring CrystalKV is accurately retained without disrupting reasoning flow. (iii) We introduce an adaptive cache budget allocation algorithm. Based on the dynamic proportion of CrystalKV, it estimates the importance of each layer/head and adjusts the KV cache budget during inference, amplifying the role of critical components to improve cache space utilization.

Results show that Crystal-KV substantially reduces memory usage by an average of 90.89\%, improves throughput by 7.57$\times$ on average, and achieves up to 1.24$\times$ speedup in user-level response latency, all while maintaining lossless compression. Even more compelling, under complex long-sequence reasoning tasks, Crystal-KV achieves 105\% of FullKV accuracy using only 10\% of the KV budget, whereas existing methods attain only 30\%–67\% accuracy. The main contributions are as follows:

\begin{itemize}
  \item By mapping answer preferences into think-stage attention map, we distinguish unified attention patterns of CrystalKV and SlipKV, laying foundation for answer-first principle.
  
  \item We propose an attention-based LRFU algorithm that retains CrystalKV while maintaining the reasoning flow.
  
  \item We introduce an adaptive cache budget allocation algorithm that amplifies the influence of critical layers and heads, improving cache space utilization.
  
  \item Results show that Crystal-KV achieves state-of-the-art KV cache compression and faster response time, while maintaining, or even improving, answer accuracy in reasoning tasks with Chain-of-Thought.
\end{itemize}

\section{Background and Motivation }
\subsection{Chain-of-Thought}

Chain-of-Thought (CoT) has emerged as a mainstream technique for enhancing the reasoning capabilities of large language models (LLMs) by explicitly modeling the human-like intermediate thinking process, enabling LLMs to solve complex tasks such as mathematics and programming~\cite{guo2025deepseek,zhang2024chain,openai2025gpt5}. Although CoT still relies on traditional autoregressive token generation for knowledge expansion and logical deduction, it introduces two key distinctions. First, it generates massive think tokens (often tens of times larger than the prompt and answer), leading to substantial KV-cache overhead that becomes the major performance bottleneck~\cite{hoffman2023training,yang2025dynamic,sui2025stop,fu2025deep,arora2025training,yeo2025demystifying}. Second, unlike normal generation tasks where all outputs are exposed to the user, the think stage remains hidden or irrelevant, and users ultimately care only about the final answer.

In summary, these two distinctions respectively highlight the necessity of think-stage KV compression and the answer-first priority in CoT reasoning.

\subsection{KV Cache Compression}
During LLM autoregressive generation, each new token must attend to all previously generated tokens, and models store their intermediate states as Key-Value (KV) cache to avoid recomputation~\cite{kwon2023efficient,zhong2024distserve,lee2024infinigen,kim2025oaken,xia2025kelle,chen2025impress}. As output sequences grow, the KV cache expands proportionally, introducing substantial memory and compute overhead, which makes KV‑cache compression (i.e., evicting redundant KV entries) essential~\cite{adnan2024keyformer,zhao2024alisa,yao2025cacheblend,pan2025instattention,quinn2025longsight}. Attention scores serve as the gold standard for evaluating the importance of KV entries~\cite{xiao2024efficient,cai2025r,hu-etal-2025-raas}. Formally, for head $h$ in layer $t$, the $i$-th token $x_i$ is projected into
$q_i, k_i, v_i$, where $(k_i, v_i)$ are appended to the existing KV cache $(K_{t,h}, V_{t,h})$. The next token $y$ is then generated by attending over the entire cached history:
\begin{gather*}
    \texttt{AttnScore} = \mathrm{softmax}(q_i K_{t,h}^\top),\ y = \texttt{AttnScore}V_{t,h}.
\end{gather*}
Here, $\text{AttnScore} \in R^{1 \times i}$ denotes the attention weights over previous $i$ cached tokens. A higher value of $\texttt{AttnScore}[0, s]$ (for $0 \leq s \leq i$) indicates that the $s$‑th token contributes more to the generation of $y$, and is therefore more critical for preserving output quality. 

By estimating near-future attention scores, existing KV compression methods fairly preserve the quality of each token generation. For instance, StreamingLLM~\cite{xiao2024efficient} assumes that upcoming tokens tend to focus on the initial and most recent tokens. H2O~\cite{zhang2023h2o} and SnapKV~\cite{li2024snapkv} approximate the attention distribution of next few tokens using score accumulation and observation windows. However, CoT-based reasoning tasks emphasize answer prioritization. The shortsighted greed induced by the uniform treatment harms the quality of answer-stage tokens generated at the end. Moreover, several recent works have attempted KV compression for CoT. Raas~\cite{hu-etal-2025-raas} identifies attention handoff patterns to facilitate high-quality think tokens. R-KV~\cite{cai2025r} observes high similarity among think-stage KV entries and performs deduplication. Unfortunately, all of them inherit a fundamental flaw of token-uniform principle, treating CoT as a special case of normal long-context generation.

In summary, these limitations motivate a comprehensive rethinking of KV cache management for CoT reasoning, guided by the answer-first principle.

\begin{figure*}[t]
  \centering
  \includegraphics[width=0.90\textwidth]{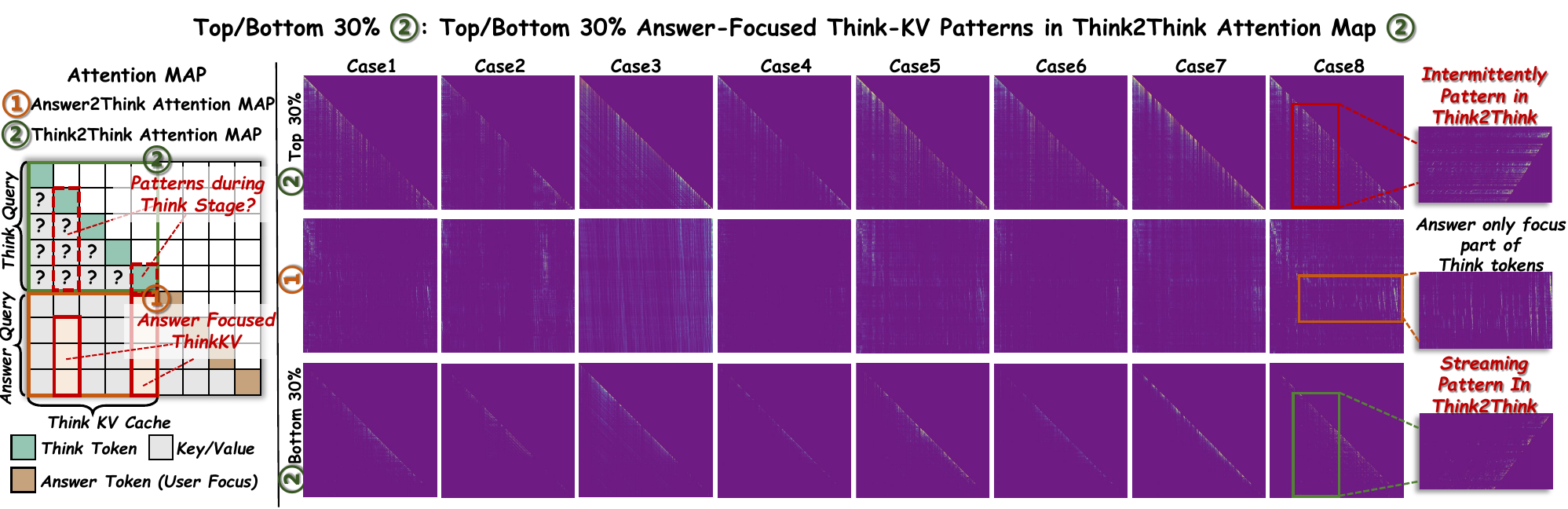}
  \caption{Distinguishing CrystalKV and SlipKV by Projecting Answer Preferences onto Think-Stage Attention Map}
  \label{fig:key_insight}
\end{figure*}
\begin{figure}[b]
  \centering
  \includegraphics[width=\linewidth]{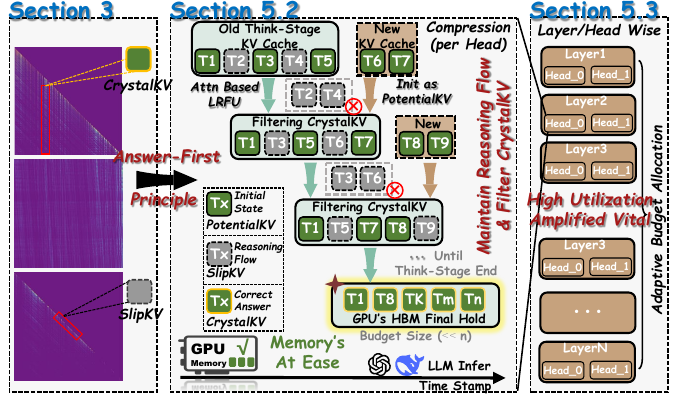}
  \caption{Overview of Crystal-KV}
  \label{fig:overview}
\end{figure}

\subsection{Budget Allocation}
KV cache budget allocation faces a trade-off between wasting memory and missing important KV entries. Pyramid~\cite{cai2024pyramidkv} introduces a layer-wise strategy: shallow layers exhibit dispersed attention and require larger budgets, while deeper layers are more focused and need less. Ada-KV~\cite{feng2025adakv} observes a similar observation across attention heads and proposes head-wise allocation. However, both rely on attention distributions of the next few tokens and are tailored for LCG compression, making them incompatible with CoT-specific methods like RaaS and R-KV. 

In summary, these limitations motivate a rethinking of how to amplify the role of key layers and heads during the think-stage KV compression, thereby improving cache space utilization.

\section{Key Insight}
\label{keyinsight}
We now revisit the question posed in Sec.~\ref{intro}: \textit{Do the think-stage KV entries that contribute to the correct answer exhibit any unified attention pattern during the think stage?} As shown in Fig.~\ref{fig:key_insight} (left), the think-stage attention map is split into two submaps \ding{172} and \ding{173}. The right panel presents attention maps from math and code reasoning tasks. Following the answer-first principle, we first examine the $(\texttt{AnswerQuery}, \texttt{ThinkKV})$ submap. As shown in the rows of right panel \ding{172}, answer tokens sparsely yet consistently attend to a specific subset of think-stage KV entries, implying that only a fraction contributes to answer generation. We then rank all think-stage KV entries by attention scores from $\texttt{AnswerQuery}$ and project top and bottom 30\% to $(\texttt{ThinkQuery}, \texttt{ThinkKV})$ submap. Top-30\% entries show a sink-to-bottom pattern, intermittent rather than continuous attention that persists until the end of the think stage, referred to as CrystalKV. Bottom-30\% entries show a streaming pattern, receiving strong local attention before gradually fading, which we call SlipKV. Functionally, CrystalKV underpins answer correctness, whereas SlipKV supports the maintenance of the ongoing reasoning process.

This two-step analysis reveals a unified attention pattern that distinguishes CrystalKV and SlipKV, serving as the foundation of the answer-first principle.

\section{Overview}
Fig.~\ref{fig:overview} presents the overview of Crystal-KV, a KV cache management framework tailored for think-stage KV compression. First, by distilling two unified attention patterns, we distinguish CrystalKV, which contributes to answer correctness and SlipKV, which maintains the reasoning flow, forming the foundation of the answer-first principle (see Sec.~\ref{keyinsight}). Then, during the think stage, an attention-based LRFU policy accurately retains CrystalKV while maintaining ongoing reasoning. Instead of explicitly identifying and locking CrystalKV, we adopt an evolutionary perspective and treat all newly arrived KV entries as PotentialKV. When a PotentialKV transitions into SlipKV, it is evicted, indicating its utility in supporting the reasoning flow expires. Therefore, only CrystalKV with persistent and answer-oriented contribution remains in the cache by the end of the think stage (see Sec.~\ref{attn_based_lrfu}). Finally, the adaptive budget allocator dynamically adjusts per-layer and per-head cache sizes, amplifying the impact of critical layer/head to improve overall cache utilization (see Sec.~\ref{adaptive_budget_allocation}). In practice, Crystal-KV achieves high answer accuracy with extremely low memory usage and fast response time (Sec.~\ref{experiment}).

\section{Implementation Details}
\subsection{Beyond Local Perspective}
In Fig.~\ref{fig:key_insight}, the attention pattern is remarkably clear from a global perspective after the think stage ends. However, during the think stage, we only have access to local attention views, making it non-trivial to distinguish CrystalKV from SlipKV. 

The key challenge arises from the local perspective: within a short time window after a new KV arrives, the boundary between CrystalKV and SlipKV is ambiguous. As shown in Fig.~\ref{fig:key_insight}, some SlipKV may have long utility spans for maintaining reasoning flow. If H2O is used, these SlipKV may be mistakenly treated as CrystalKV, occupying valuable cache space. If StreamingLLM is used, these SlipKV may be evicted too early, disrupting reasoning process. Moreover, CrystalKV tends to receive intermittent attention. If SnapKV is used, it may be misclassified as SlipKV and evicted during attention gaps.

These limitations motivate us to distinguish CrystalKV and SlipKV from a temporal and evolutionary perspective. Specifically, we initialize all newly arrived KV entries as PotentialKV. During think stage, each PotentialKV may evolve into CrystalKV (and be retained) or SlipKV (and be evicted). However, due to the presence of attention gaps, the transition from PotentialKV to CrystalKV is not intuitive. Shifting our focus, however, we identify a qualitative criterion for detecting the transition from PotentialKV to SlipKV: at the current time step, if a PotentialKV is attended infrequently in its past history and remains unattended for a sufficiently long recent period, it can be safely classified as SlipKV and evicted. By prioritizing the identification and eviction of SlipKV, the cache is left with only CrystalKV, ensuring correct answers. Fortunately, when attempting to quantify this criterion, we find that it closely aligns with the classical cache replacement policy: Least Recently Frequently Used (LRFU)~\cite{lee2001lrfu}.

\begin{algorithm}[t]
\caption{Attention-Based LRFU Compression (per Head)}
\label{alg:lrfu-per-head}
\small
\renewcommand{\baselinestretch}{1.10}\selectfont
\begin{algorithmic}[1]
\Statex\hspace{-\algorithmicindent} \hspace*{-0.5em} \textbf{Inputs:} 
$\texttt{B}$: KV budget, 
$\texttt{p}$: top-$p$ threshold for hit mask, 
$\lambda$: CRF decay rate, 
$\texttt{t}$: current time step, 
$\texttt{K}/\texttt{V}$: key/value cache (PotentialKV), 
$\texttt{Q}^{\texttt{last}}$: last query state, 
$\texttt{CRF}$: CRF score for all KV, 
$\tau$: last update time for all KV.

\Statex\hspace{-\algorithmicindent} \hspace*{-0.5em} \textbf{Outputs:} 
$\texttt{K}'/\texttt{V}'$: compressed key/value cache.

\State \textbf{if} $Len(\texttt{K}) < \texttt{B}$ \textbf{then}
\State \hspace*{0.5em} \textbf{return} $\texttt{K}$,\ $\texttt{V}$
\cmt{Step 1: 0-1 Hit KV Mask via Top-$p$ Attention}
\State $\texttt{s} \gets \textsc{AttnScores}(\texttt{Q}^{\texttt{last}},\ \texttt{K})$
\State $\texttt{M}_{hit} \gets \textsc{GetHitMask}(\texttt{s},\ \texttt{p})$ 
\cmt{Step 2: Update and Record Hit KV CRF}
\State $\texttt{CRF}_{temp}\texttt{[}\texttt{M}_{hit}\texttt{]}\gets \lambda^{t - \tau\texttt{[} \texttt{M}_{hit}\texttt{]}} \cdot \texttt{CRF[}\texttt{M}_{hit}\texttt{]} + 1$
\State $\tau\texttt{[}\texttt{M}_{hit}\texttt{]} = \texttt{t}$
\State $\texttt{CRF}\texttt{[}\texttt{M}_{hit}\texttt{]} \gets \texttt{CRF}_{temp}\texttt{[}\texttt{M}_{hit}\texttt{]}$
\cmt{Step 3: Update Miss KV CRF without Record}
\State $\texttt{M}_{miss}\gets 1 - \texttt{M}_{hit}$ 
\State $\texttt{CRF}_{temp}\texttt{[}\texttt{M}_{miss}\texttt{]}\gets \lambda^{t - \tau\texttt{[} \texttt{M}_{miss}\texttt{]}} \cdot \texttt{CRF[}\texttt{M}_{miss}\texttt{]}$
\cmt{Step 4: SlipKV Identification and Eviction}
\State $\texttt{Retain} \gets \textsc{TopK}(\texttt{CRF}_{temp},\ \texttt{B}).\texttt{index}$
\State $\texttt{CRF} \gets \textsc{Gather}(\texttt{CRF},\ \texttt{Retain})$, $\tau \gets \textsc{Gather}(\tau,\ \texttt{Retain})$ 
\State $\texttt{K}' \gets \textsc{Gather}(\texttt{K},\ \texttt{Retain})$, $\texttt{V}' \gets \textsc{Gather}(\texttt{V},\ \texttt{Retain})$ 

\State \textbf{return} $\texttt{K}',\ \texttt{V}'$
\end{algorithmic}
\end{algorithm}

\subsection{Attention Based LRFU}
\label{attn_based_lrfu}
Along the temporal dimension, LRFU integrates both access (hit) frequency and recency to assess the importance of each entry using the Combined Recency and Frequency (CRF) score. Given current time $t$, for an entry $e_i$ with historical hit timestamps ${t_1, t_2, \ldots, t_k}$ and decay factor $\lambda \in [0, 1]$, the CRF score is defined as:
\begin{gather*}
CRF(e_{i}) = \sum_{j=1}^{k} \lambda^{t-t_j}.
\end{gather*}
Intuitively, when an entry $e_i$ is hit at time $t_j$, it receives a reward of $\lambda^{t_j - t_j} = 1$, which then decays over time to $\lambda^{t - t_j}$ at a later time $t$. When the cache is full, entries with smaller CRF scores are evicted.

The CRF-based eviction policy precisely models the transition from PotentialKV to SlipKV. However, we cannot observe hits as in CPU instructions directly. Thus, we perform Top‑$p$ nucleus sampling based on attention scores, selecting the smallest set of KV whose cumulative attention scores exceed a threshold $p$ as hit KV entries. To reduce memory overhead, we further implement incremental CRF tracking. For each $e_{i}$, we only store its last hit time $t_{i}$, and the CRF score is updated at any time $t$ as:
\begin{gather*}
    CRF_i(t) = \lambda^{t-t_i} \cdot CRF_i(t_i) + 1_{\{\texttt{hit}\ \texttt{at}\ t\}}
\end{gather*}
Algorithm~\ref{alg:lrfu-per-head} outlines the KV cache compression per head. First, we apply Top‑$p$ nucleus sampling on attention scores to derive a boolean matrix $\texttt{M}_{\texttt{hit}}$, indicating which entries are hits (lines 3–4). For hits, we update and record both CRF and last-hit time (lines 5-7). For misses, we only update CRF (lines 8-9). Finally, entries with the lowest CRF are classified as SlipKV and evicted accordingly (lines 10–13). Throughout the think stage, it gradually compresses the KV cache by dropping SlipKV, ensuring only CrystalKV remains before entering the answer stage.

\begin{algorithm}[t]
\caption{Adaptive Budget Allocation (Layer\ \&\ Head)}
\label{alg:two-level-scheduler}
\small
\renewcommand{\baselinestretch}{1.10}\selectfont
\begin{algorithmic}[1]
\Statex\hspace{-\algorithmicindent} \hspace*{-0.7em} \textbf{Inputs:} 
$\texttt{B}_{total}$: total KV budget for LLM, 
$\texttt{L}$: layer num of LLM,
$\texttt{H}$: head num per layer,
$\texttt{CRF}_{i,j}$: aggregated CRF scores of head j($<\texttt{H}$) in layer i($<\texttt{L}$), 
${\texttt{B}_{i,j}}$: current KV budget for head j($<\texttt{H}$) in layer i($<\texttt{L}$)
\Statex\hspace{-\algorithmicindent} \hspace*{-0.7em} \textbf{Outputs:} 
${\texttt{B}'_{i,j}}$: new KV budget for head j($<\texttt{H}$) in layer i($<\texttt{L}$)
\cmt{Step 1: Budget Utilization per Layer}
\State $\eta_{i} \gets \sum_{j=1}^{h}\texttt{CRF}_{i,j}\ /\ \sum_{j=1}^{h}\texttt{B}_{i,j}$
\cmt{Step 2: Layer-wise Budget Allocation}
\State $\texttt{B}'_{i} \gets \texttt{B}_{total} \cdot (\eta_{i}\ /\ \sum_{i=1}^{l}\eta_{i})$
\cmt{Step 3: Budget Utilization per Head}
\State $\eta_{i,j} \gets \texttt{CRF}_{i,j} \ /\ \texttt{B}_{i,j}$
\cmt{Step 4: Head-wise Budget Allocation}
\State $\texttt{B}'_{i,j} \gets \texttt{B}'_{i} \cdot (\eta_{i,j}\ /\ \sum_{j=1}^{h}\eta_{i,j})$

\State \Return $\texttt{B}'_{i,j}$
\end{algorithmic}
\end{algorithm}

Notably, when $\lambda = 0$, the algorithm degenerates to Least Recently Used (LRU)~\cite{belady1966study}, focusing only on recency. When $\lambda = 1$, it becomes Least Frequently Used (LFU)~\cite{wiki_lfu}, focusing purely on frequency. Therefore, we constrain $\alpha < \lambda < \beta$. $\alpha < \lambda$ prevents excessive reward decay of CrystalKV during attention gaps, avoiding its misclassification as SlipKV. $\lambda < \beta$ ensures that long-lived SlipKV entries are not over-preserved, preventing occupation of valuable cache space. The choice of $\alpha$ and $\beta$ is further discussed in Sec.~\ref{more_disscussions}.

\subsection{Adaptive Budget Allocation}
\label{adaptive_budget_allocation}
Under a uniform and fixed budget allocation, we observe that after Algorithm~\ref{alg:lrfu-per-head}, certain layers/heads end with relatively high CRF scores, indicating that the tight budget forced them to evict CrystalKV entries important for the final answer. Conversely, some layers or heads show consistently low CRF scores, suggesting little or no CrystalKV in those regions. This reveals that different layers/heads contribute heterogeneously to the reasoning process. To capture the relative importance of each layer/head under a limited budget, we define a metric of cache utilization:
\begin{gather*}
\eta = \frac{\sum \texttt{CRF}}{\sum \texttt{Budget}}
\end{gather*}
A higher $\eta$ indicates more potential CrystalKV entries under a tighter budget, implying the layer/head is more important and should be allocated more budget. 

\begin{figure*}[t]
  \centering
  \includegraphics[width=0.81\textwidth]{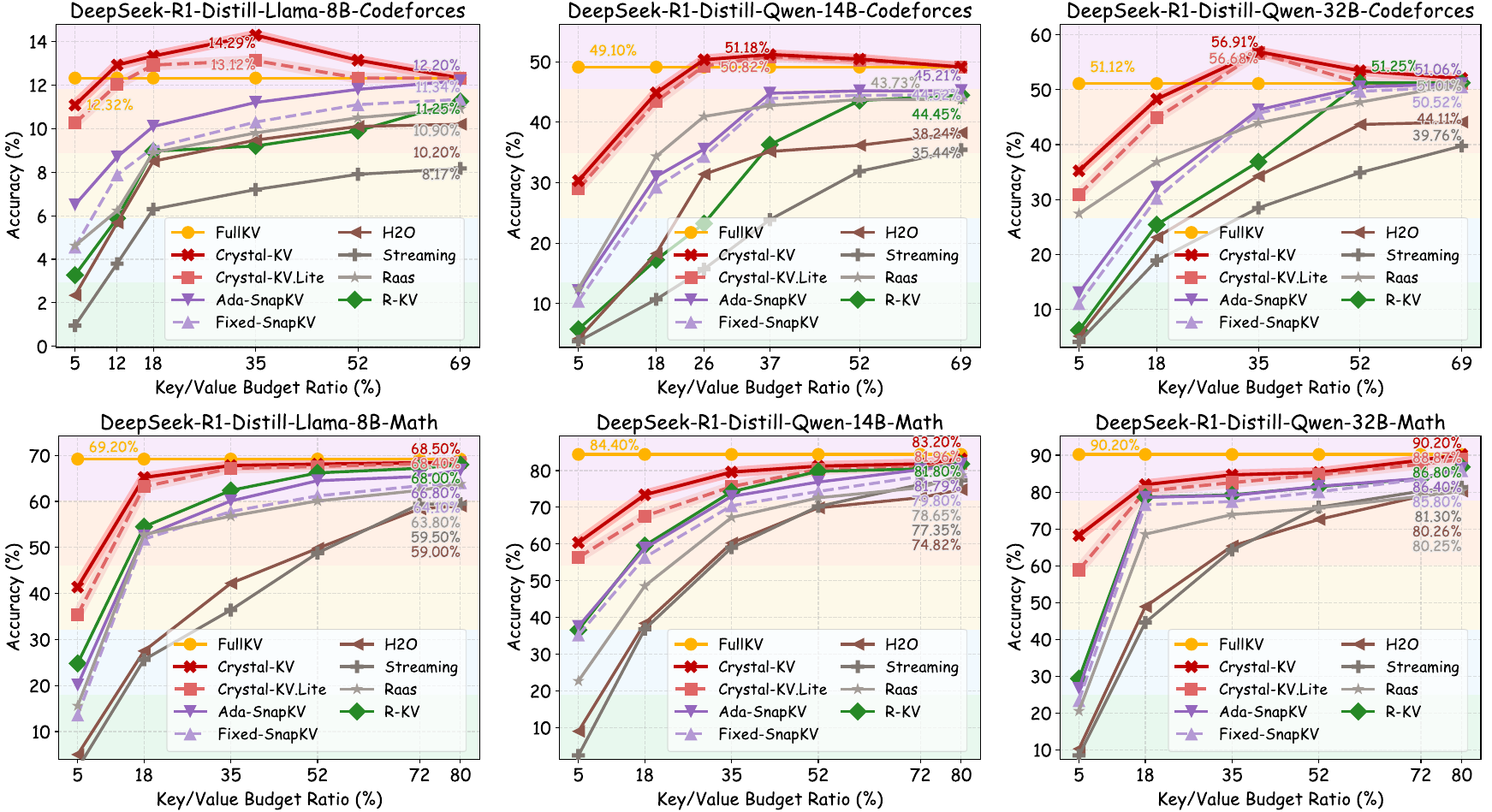}
  \caption{Accuracy Comparison on Math and Code Tasks across Three Reasoning LLMs}
  \label{fig:zhexiantu}
\end{figure*}

Algorithm~\ref{alg:two-level-scheduler} outlines the complete adaptive budget allocation process. First, it computes layer-wise utilization by aggregating CRF scores and budgets across all heads (line 1). Then, it normalizes utilization across layers and adjusts their budget proportionally (line 2). The same procedure is applied to heads within each layer (lines 3–4). This algorithm is periodically triggered. By amplifying the impact of critical layers/heads, adaptive budget allocation significantly improves cache space efficiency.

\section{Experiment}
\label{experiment}
\subsection{Experiment Setup}
\textbf{Models and Datasets}. We evaluate three DeepSeek-R1–distilled open-source models (Llama-8B, Qwen-14B, and Qwen-32B)~\cite{guo2025deepseek}, chosen for their strong CoT reasoning performance. Our experiments span two challenging reasoning domains: programming using the CodeForces benchmark (10K competitive programming problems up to 2025)~\cite{penedo2025codeforces} and mathematics using MATH-500 (advanced competition-level math problems)~\cite{hendrycks2021measuring}.

\noindent \textbf{Baselines}. We compare Crystal-KV with five representative KV compression baselines: R-KV, RaaS, SnapKV, StreamingLLM, and H2O. SnapKV, StreamingLLM, and H2O are long-context generation (LCG) compress methods that follow the token-uniform principle, while R-KV and RaaS are CoT-oriented compressors but still treat CoT as a special case of LCG. Since all these baselines adopt static, uniform, and fixed budget allocation, we introduce Crystal-KV.Lite, a variant of our method where dynamic budget allocation is disabled, to ensure a fair comparison. Moreover, we include Ada-SnapKV, which augments SnapKV with dynamic budget reallocation, enabling a direct comparison with the full-featured Crystal-KV. It is worth noting that due to inherent limitations in their compression strategies, other baselines cannot be adapted to support dynamic reallocation. Finally, we include FullKV, which retains the full cache and serves as the gold standard.

\begin{table*}[t]
\renewcommand\arraystretch{0.6}
\setlength\tabcolsep{4pt}
\centering
\caption{Comparison of Memory Consumption and Inference Throughput}
\resizebox{\textwidth}{!}{
\begin{tabular}{cccccccc}
\toprule
\huge \textbf{Gen.Len.} & \huge \textbf{Method} & \huge \textbf{Budget} & \huge \textbf{HBM Saving (\%)} & \huge \textbf{Batch} & \huge \textbf{Throughput (tok/s)} & \huge \textbf{Tokens Gen.} & \huge \textbf{Dec.Time (s)} \\
\midrule

\multirow{5}{*}{{\Huge 8K-level}}
& \multirow{2}{*}{{\Huge FullKV}}
& {\Huge -} & {\Huge -} & {\Huge 1} & {\Huge 53.14} & {\Huge 9,350} & {\Huge 175.95} \\
& & {\Huge -} & {\Huge -} & {\Huge 51 (max)} & {\Huge 458.67} & {\Huge 476,850 (max)} & {\Huge 1039.63} \\
\cmidrule(lr){2-8}

& \multirow{3}{*}{{\Huge Crystal-KV} }
& \cellcolor[gray]{0.95}{\Huge Fixed-1024} & \cellcolor[gray]{0.95}{\Huge 89.05} & \cellcolor[gray]{0.95}{\Huge 1} & \cellcolor[gray]{0.95}{\Huge 59.58} & \cellcolor[gray]{0.95}{\Huge 9,350} & \cellcolor[gray]{0.95}{\Huge 156.93} \\
& & \cellcolor[gray]{0.95}{\Huge Fixed-1024} & \cellcolor[gray]{0.95}{\Huge 89.05} & \cellcolor[gray]{0.95}{\Huge 379 (max)} & \cellcolor[gray]{0.95}{\Huge 2297.37} & \cellcolor[gray]{0.95}{\Huge 3,543,650 (max)} & \cellcolor[gray]{0.95}{\Huge 1542.48} \\
& & \cellcolor[gray]{0.95}{\Huge Ratio-10\%-935} & \cellcolor[gray]{0.95}{\Huge 90.00} & \cellcolor[gray]{0.95}{\Huge 412 (max)} & \cellcolor[gray]{0.95}{\Huge 2478.39} & \cellcolor[gray]{0.95}{\Huge 3,852,200 (max)} & \cellcolor[gray]{0.95}{\Huge 1554.31} \\

\midrule

\multirow{5}{*}{{\Huge 16K-level}}
& \multirow{2}{*}{{\Huge FullKV}}
& {\Huge -} & {\Huge -} & {\Huge 1} & {\Huge 41.99} & {\Huge 18,700} & {\Huge 445.34} \\
& & {\Huge -} & {\Huge -} & {\Huge 25 (max)} & {\Huge 174.95} & {\Huge 467,500 (max)} & {\Huge 2672.22} \\
\cmidrule(lr){2-8}

& \multirow{3}{*}{{\Huge Crystal-KV}}
& \cellcolor[gray]{0.95}{\Huge Fixed-1024} & \cellcolor[gray]{0.95}{\Huge 94.52} & \cellcolor[gray]{0.95}{\Huge 1} & \cellcolor[gray]{0.95}{\Huge 56.90} & \cellcolor[gray]{0.95}{\Huge 18,700} & \cellcolor[gray]{0.95}{\Huge 328.65} \\
& & \cellcolor[gray]{0.95}{\Huge Fixed-1024} & \cellcolor[gray]{0.95}{\Huge 94.52} & \cellcolor[gray]{0.95}{\Huge 379 (max)} & \cellcolor[gray]{0.95}{\Huge 2141.44} & \cellcolor[gray]{0.95}{\Huge 7,087,300 (max)} & \cellcolor[gray]{0.95}{\Huge 3309.59} \\
& & \cellcolor[gray]{0.95}{\Huge Ratio-10\%-1870} & \cellcolor[gray]{0.95}{\Huge 90.00} & \cellcolor[gray]{0.95}{\Huge 228 (max)} & \cellcolor[gray]{0.95}{\Huge 1332.69} & \cellcolor[gray]{0.95}{\Huge 4,263,600 (max)} & \cellcolor[gray]{0.95}{\Huge 3199.25} \\

\bottomrule
\end{tabular}
}
\label{throughput}
\end{table*}

\noindent \textbf{Evaluation Settings}. Due to the limited ability of LLMs on challenging coding tasks, we restrict our CodeForces evaluation to problems with difficulty ratings below 1500~\cite{deepseekai2025deepseekr1,guo2025deepseek}, to avoid accuracy saturation or complete failure, ensuring headroom for performance improvements. We use the recommended sampling temperature (0.6) and top-p (0.95)~\cite{guo2025deepseek}. For each problem, we generate $k=8$ answers, and report the average correctness scores, computed as $ \frac{1}{k} \sum_{i=1}^{k} p_i$, where $p_i$ denotes the correctness of the $i$-th answer. This procedure yields more reliable accuracy estimates. All experiments are conducted on a workstation equipped with three NVIDIA RTX PRO 6000 Blackwell GPUs.

\subsection{Accuracy Comparison}
Fig.~\ref{fig:zhexiantu} reports the accuracy of all methods across different tasks and KV-budgets. Budget Ratio is defined relative to the average per-question KV usage of FullKV. We can distill four key observations: (i) Under the same KV budget, Crystal-KV consistently outperforms StreamingLLM, H2O, R-KV, Fixed-SnapKV, Ada-SnapKV, and RaaS. On code tasks, the average accuracy gains are 18.98\%, 14.36\%, 12.46\%, 9.04\%, 7.91\%, and 7.30\%. On math tasks, the average accuracy gains are 23.87\%, 23.05\%, 7.97\%, 11.39\%, 9.13\%, and 14.65\%. (ii) Focusing more attention on CrystalKV during the answer stage can turn previously wrong answers into correct ones. On code tasks, a peak appears where Crystal-KV even surpasses FullKV. Before this peak, the budget is too small to retain all CrystalKV, so increasing budget improves accuracy. After this point, the KV budget exceeds the total amount of CrystalKV, causing SlipKV entries to be stored as well, so the accuracy gradually converges back to the FullKV level. (iii) Adaptive Budget Allocation further improves performance under very small budgets by enhancing space utilization and amplifying critical layers and heads. (iv) Accuracy gains on math are smaller than on code under the same budget. This is because math tasks contain more dispersed key information, requiring larger CrystalKV budgets to match the performance level.

\subsection{Efficient Memory Saving and Computation}
Under the FullKV strategy, GPU memory usage grows linearly with the reasoning length, making long-sequence reasoning prone to Out of Memory (OOM), and the cost of computation increases. In contrast, Crystal-KV enforces a fixed memory budget, significantly reducing both memory usage and computation cost.

Table~\ref{throughput} compares Crystal-KV with the FullKV baseline on long-sequence reasoning using DeepSeek-R1-Distill-Llama-8B with a total HBM capacity of 288 GB. The average reasoning lengths are 9,350 and 18,700 tokens (8K and 16K levels), consistent with practical CodeForces usage. We evaluate both fixed and ratio-based budgets to accommodate both strict and flexible memory constraints. Existing KV-compression baselines are not considered here because they cannot attain FullKV-level accuracy under strict budgets in long-sequence reasoning. At 8K level, Crystal-KV accelerates single-batch inference (response time) by 6.44 token/s over FullKV. Given the total GPU HBM capacity, FullKV can support at most 51 parallel batches. In contrast, Crystal-KV can support 379 parallel batches under a fixed budget of 1024 KV entries per head, and 412 batches under a 10\% ratioed budget (a setting that already achieves lossless compression). The corresponding throughput is improved by 5.01$\times$ and 5.40$\times$ over FullKV, and maximum total number of tokens that can be processed concurrently reaches 7.43$\times$ and 8.07$\times$ that of FullKV. When the reasoning problem becomes more difficult and the sequence length increases to the 16K level, the advantage of Crystal-KV becomes even more pronounced. The overall throughput is boosted to 12.24$\times$ and 7.62$\times$, and the maximum total number of tokens increases to 15.16$\times$ and 9.12$\times$. This indicates that, in scenarios requiring much longer reasoning chains (i.e., more complex reasoning tasks), Crystal-KV has substantial potential to handle problems that are infeasible for FullKV, owing to its ability to handle much longer reasoning chains.

In summary, Crystal-KV substantially reduces memory usage by an average of 90.89\%, improves throughput by 7.57$\times$ on average, and achieves up to 1.24$\times$ speedup in user-level response latency, all while maintaining lossless compression.
\begin{figure}[t]
  \centering
  \includegraphics[width=1\linewidth]{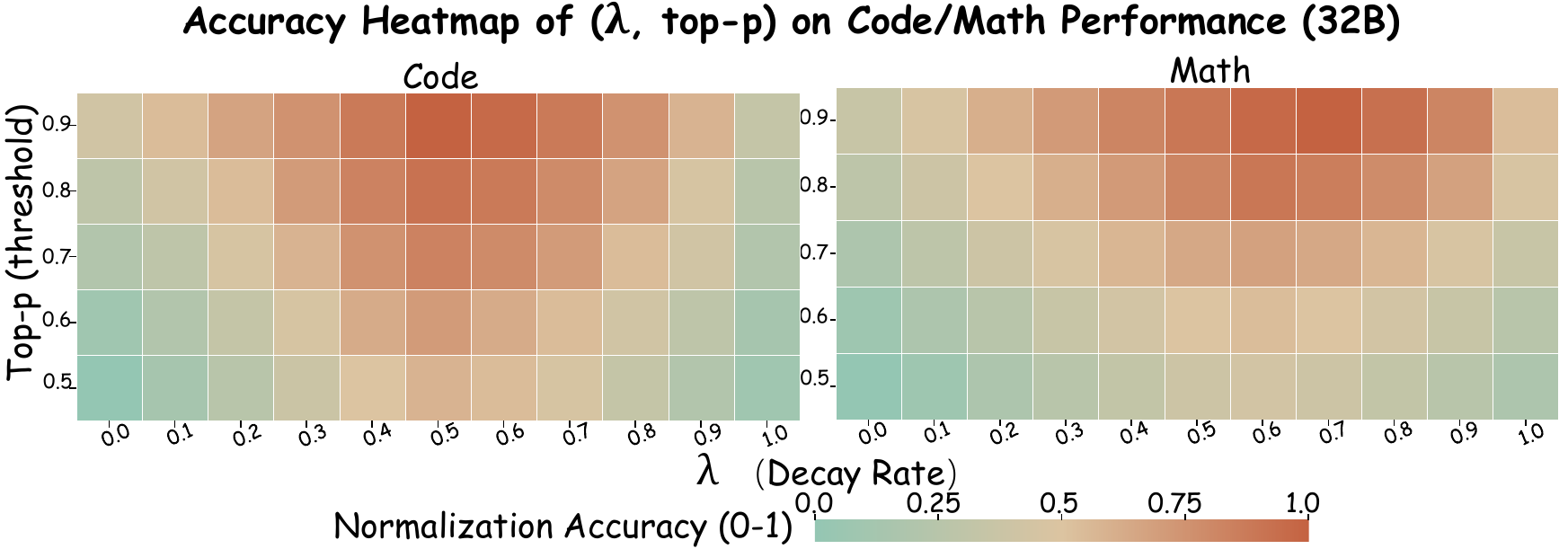}
  \caption{Effect of Decay Rate and Sampling Threshold}
  \Description{A heatmap showing how $\lambda$ and top-p.}
  \label{fig:ablation}
\end{figure}

\subsection{More Discussions}
\label{more_disscussions}
Fig.~\ref{fig:ablation} shows accuracy heatmaps of Crystal-KV on Codeforces and MATH500 under different sampling thresholds ($top\texttt{-}p \in \{0.5$, $0.6$, $0.7$, $0.8$, $0.9$) and decay factors ($\lambda \in \{0.0,0.1,\ldots,1.0\}$). Both tasks exhibit unified patterns. For $\lambda$, accuracy peaks at $[0.6,0.7]$ for code and $[0.5,0.6]$ for math. When $\lambda$ is too small, historical rewards for CrystalKV decay excessively during attention gaps, causing misclassification as SlipKV. When $\lambda$ is too large, overemphasis on historical hits delays the eviction of long-lived SlipKV and occupies space needed for CrystalKV. Furthermore, accuracy for $\lambda \in [0.8,1.0]$ exceeds that for $[0,0.2]$, suggesting that most CrystalKV entries undergo long attention gaps while SlipKV entries have short lifetimes. For $top\texttt{-}p$, higher thresholds (e.g., 0.9) better preserve CrystalKV and prevent the loss of critical context, whereas lower thresholds (e.g., 0.5) produce volatile attention patterns and make $\lambda$ more sensitive. Thus, we recommend $top$-$p = 0.9$ for robust performance and set $\lambda$ within a moderate range, approximately $[0.5, 0.7]$, corresponding to the $(\alpha, \beta)$ bounds discussed in Sec.~\ref{attn_based_lrfu}.

\section{Conclusion}
In this work, we present Crystal-KV, an efficient CoT KV-cache management framework built on an answer-first principle that prioritizes think-KV essential to the final answer. This perspective redefines KV management for reasoning-oriented LLMs during long think stage. Experiments show that Crystal-KV delivers substantial memory savings and latency improvements while maintaining or even improving answer accuracy across diverse reasoning tasks. These findings establish answer-centric cache management as a promising paradigm for efficient large-scale reasoning. In future work, we will formally model CrystalKV and SlipKV to quantify the capability boundary of KV-cache compression for CoT reasoning.

\newpage

\newpage
\bibliographystyle{ACM-Reference-Format}
\bibliography{ref}

\end{document}